\documentclass[twocolumn,twoside]{IEEEtran}
\usepackage{cite}
\usepackage{graphicx}			
\usepackage{color}
\usepackage{float}
\usepackage{wrapfig}
\usepackage{amsmath,mathrsfs,bm}
\usepackage{graphicx}			
\usepackage{tabularx,ragged2e,booktabs,caption}
\usepackage{subcaption}
\usepackage{comment}


\title{Sparse Reconstruction of Compressive Sensing MRI using Cross-Domain Stochastically Fully Connected Conditional Random Fields}
\author{Edward Li, Farzad Khalvati$^*$, Mohammad Javad Shafiee, Masoom A. Haider, Alexander Wong
\thanks{E. Li, M. J. Shafiee, and A. Wong are with the Department of Systems Design Engineering, University of Waterloo, Waterloo, Ontario, Canada, N2L 3G1. {\tt\small a28wong@uwaterloo.ca}}
\thanks{F. Khalvati$^*$ and M. A. Haider are with the Department of Medical Imaging, University of Toronto and Sunnybrook Research Institute, Toronto, Ontario, Canada, M4N 3M5. {\tt\small farzad.khalvati@sri.utoronto.ca}}

}

\begin{document}

\twocolumn

\maketitle

\begin{abstract} - Magnetic Resonance Imaging (MRI) is a crucial medical imaging technology for the screening and diagnosis of frequently occurring cancers. However image quality may suffer by long acquisition times for MRIs due to patient motion, as well as result in great patient discomfort. Reducing MRI acquisition time can reduce patient discomfort and as a result reduces motion artifacts from the acquisition process. Compressive sensing strategies, when applied to MRI, have been demonstrated to be effective at decreasing acquisition times significantly by sparsely sampling the \emph{k}-space during the acquisition process. However, such a strategy requires advanced reconstruction algorithms to produce high quality and reliable images from compressive sensing MRI. This paper proposes a new reconstruction approach based on cross-domain stochastically fully connected conditional random fields (CD-SFCRF) for compressive sensing MRI. The CD-SFCRF introduces constraints in both \emph{k}-space and spatial domains within a stochastically fully connected graphical model to produce improved MRI reconstruction. Experimental results using T2-weighted (T2w) imaging and diffusion-weighted imaging (DWI) of the prostate show strong performance in preserving fine details and tissue structures in the reconstructed images when compared to other tested methods even at low sampling rates.
\end{abstract}

\section{Introduction}
Magnetic Resonance Imaging (MRI) is a medical imaging technology that is currently used for diagnostic imaging of a wide range of diseases. In particular, since MRI does not use ionizing radiation, it has been becoming a crucial imaging modality for screening frequently occurring cancers such as prostate cancer in men, breast cancer in women, as well as lung and colorectal cancer for both men and women. In 2015, 196,900 new cases of cancer (excluding non-melanoma skin cancers) were expected, with 51\% of these belonging to the four aforementioned types of cancer in Canada~\cite{CCancerStats}. As such, cancer screening methods with accurate and reliable information such as MRI is highly desired.  Of particularly increasing interest for cancer screening is multi-parametric MRI (MP-MRI) since more information can be acquired through different modalities. MP-MRI contains different modalities such as diffusion weighted imaging (DWI), correlated diffusion imaging (CDI)~\cite{WongCDI,Khalvati2015_1}, dynamic contrast enhancement (DCE), T2-weighted (T2w) imaging, and T1-weighted (T1w) imaging~\cite{esen2014}. Although this approach provides a more complete information, acquisition times are significantly longer which causes higher patient discomfort and motion artifacts that decreases image quality. Due to this fact, new methods to improve MRI acquisition time are highly desired to facilitate for reliable MP-MRI data acquisition.

Compressive sensing has demonstrated to be an effective strategy for reducing MRI acquisition times by acquiring significantly fewer samples in \emph{k}-space. A complete signal can be then be reconstructed fully through sparse, yet sufficient number of samples~\cite{Donoho2006,candes2006,Baraniuk2007}. In MRI, compressive sampling strategies have been demonstrated to be highly effective at reducing acquisition time while maintaining image quality as different types of tissue structure have been shown to be sparse in certain domains~\cite{Lustig2007}. Furthermore, different techniques have been proposed to improve the imaging process~\cite{Liang2009} as well as the reconstruction process~\cite{Ye2007,Wong2013,Qu2010,Block2007,trzasko2009,chartrand2009fast,yin2008bregman,figueiredo2007gradient,goldstein2009split,wang2008new,osher2005iterative,Qu2008,Yu2011} in compressive sensing. Due to the limited amount of data available through compressive sensing, advanced reconstruction algorithms are required to produce high quality reliable images which the ongoing challenges mainly span in improving the reconstruction algorithms, efficiency and quality of compressive sensing MRI.

A number of different methods have been proposed for sparse reconstruction of compressive sensing MRI~\cite{Ye2007,Wong2013,Qu2010,Block2007,trzasko2009,chartrand2009fast,yin2008bregman,figueiredo2007gradient,goldstein2009split,wang2008new,osher2005iterative,Qu2008,Yu2011}.  As a notable example, Block \emph{et al.} \cite{Block2007} proposed an iterative image reconstruction technique using a modified total variation (TV)~\cite{wang2008new,osher2005iterative} constraint for sparse reconstruction of compressive sensing brain MRI.  Trzasko \emph{et al.} \cite{trzasko2009} introduced a homotopic \emph{l$_{0}$} minimization method for the sparse reconstruction of compressive sensing spinal MRI.  Wong \emph{et al.}~\cite{Wong2013} extended upon this idea to a regional sparsified domain for the sparse reconstruction of breast MRI.  A similar technique was also demonstrated by Qu \emph{et al.} using combined sparsifying transforms and smoothed \emph{l$_{0}$} norm minimization \cite{Qu2010}, where they showed that the use of combined transforms can improve image quality compared of the reconstructed images from compressive sensing MRI when compared to methods using a single sparsifying transform.

An area that is little explored but can reap significant potential benefits is the application of random field modeling for improved sparse reconstruction of compressive sensing MRI.  Random field modeling such as Markov random fields (MRF) \cite{held1997markov,blake2011markov} and conditional random fields (CRF) \cite{lafferty2001conditional} have long been shown to be powerful tools for incorporating spatial context within a probabilistic graphical modeling framework, which can have significant benefits for reconstructing images from sparse measurements.  Despite its powerful modeling capabilities and potential benefit to sparse reconstruction, one of the biggest hurdles in leveraging random field models for compressive sensing MRI is the fact that all MRI measurements are made in \emph{k}-space, whereas the reconstructed image exists in the spatial domain.  As the majority of random field models typically model in a single domain, such models cannot be used directly for the purpose of sparse reconstruction of compressive sensing MRI.  This is further complicated by the fact that the MRI measurements in \emph{k}-space are sparse and incomplete, which makes it difficult to leverage existing random field models for this problem.  Therefore, a probabilistic graphical modeling framework that can consolidate the fact that partial measurements are made in a domain different than the desired states of the reconstruction images is needed to truly leverage the power of random field modeling for sparse reconstruction of compressed sensing MRI.

This paper proposes a cross-domain Stochastically fully connected conditional random field (CD-SFCRF) approach for the reconstruction of compressive sensing MRI at below Nyquist sampling rates \cite{Nyquist1928}. The CD-SFCRF framework introduces constraints in both \emph{k}-space and spatial domains within a stochastically fully connected graphical model~\cite{Shafiee2013} to produce improved MRI reconstruction. The proposed CD-SFCRF framework has the ability to utilize spatial and data driven consistencies in the spatial domain along with data driven consistencies in the \emph{k}-space domain pertaining to sparse measurements while maintaining edge features and structural details in the reconstructed images. Phantom MRI data as well as prostate MRI data captured using T2w and DWI imaging modalities, which also yields apparent diffusion coefficient (ADC) map images, are used to illustrate the efficacy of the proposed CD-SFCRF framework for sparse reconstruction of compressive sensing MRI. To the best of the authors' knowledge, this is the first time that constraints in both \emph{k}-space and spatial domains are used in conjunction within a stochastically fully connected graphical model for the sparse reconstruction of compressive sensing MRI, which is the main contribution of this paper.

The paper is formatted as follows. The methodology behind the proposed CD-SFCRF framework is described in Section II. The Experimental setup is described in Section III. Results and comparisons with previous methods are discussed in Section IV. Finally, conclusions will be drawn and future work will be discussed in Section V.

\section{Methodology}

In MRI, measurements are made in the \emph{k}-space~\cite{ljunggren1983simple}, with the lower frequency coefficients in the \emph{k}-space containing coarse-grained contrast information while higher frequency coefficients contain fine-grained image detail information.  The MRI measurements from the \emph{k}-space are transformed into the spatial domain to form the reconstructed MRI image.  Most compressive sensing strategies \cite{Donoho2006,trzasko2009} sparsely sample the \emph{k}-space to reduce image acquisition time significantly. Therefore, to fully utilize available information in the reconstruction process, data-driven constraints in the \emph{k}-space domain and data and spatial driven constraints in the spatial domain would be highly beneficial in improving image reconstruction quality from compressive sensing MRI.

Motivated by this, the proposed cross-domain stochastically fully connected conditional random field (CD-SFCRF), introduced here for the purpose of sparse reconstruction of compressive sensing MRI, extends upon the seminal work on stochastically fully connected conditional random fields (SFCRF) first proposed in~\cite{Shafiee2013} to facilitate for this cross-domain optimization.  Let us first discuss the concept of SFCRFs briefly for context upon which we build CD-SFCRF upon.  SFCRFs are fully-connected conditional random fields with stochastically defined cliques. Unlike traditional conditional random fields (CRF) where nodal interactions are deterministic and restricted to local neighborhoods, each node in the graph representing a SFCRF is connected to every other node in the graph, with the cliques for each node is stochastically determined based on a distribution probability. Therefore, the number of pairwise cliques might not be the same as the number of neighborhood pairs as in the traditional CRF models.  By leveraging long-range nodal interactions in a stochastic manner, SFCRFs facilitate for improved detail preservation while maintaining similar computational complexity as CRFs, which makes SFCRFs particularly enticing for the purpose of improved sparse reconstruction of compressive sensing MRI.  However, here the problem is to reconstruct an MRI image in the spatial domain while the available measurements are made in \emph{k}-space domain.  Like most CRF models, SFCRFs cannot be leveraged directly for this purpose.  Motivated by the significant potential benefits of using SFCRFs in improving reconstruction quality of compressive sensing MRI, we extend the SFCRF model into a cross-domain stochastically fully connected conditional random field (CD-SFCRF) model that incorporates cross-domain information and constraints from \emph{k}-space and spatial domains to reconstruct the desirable MRI image from sparse observations in \emph{k}-space.

The main goal here is to reconstruct image $Y$ given original sparsely sampled \emph{k}-space observations $X$.  We model the conditional probability $P(Y|X)$ of the full state set $Y$ in spatial domain given the set of sparse measurements $X$ in \emph{k}-space, which can be written as:
\begin{align}
P(Y | X) = \frac{1}{Z(X)}\exp(-\psi(Y | X))
\label{ConProbEq}
\end{align}
where $Z(X)$ is the normalization function and $\psi(.)$ is a combination of unary and pairwise potential functions:
\begin{align}
\psi(Y|X) = \sum^n_{i=1}\psi_u(y_i,X) + \sum_{\varphi\in{C}}\psi_p(y_{\varphi},X)
\label{PotentialEq}
\end{align}
\noindent Here $y_i\in{Y}$ is a single state in the set $Y=\{y_i\}^n_{i=1}$, $y_{\varphi}\in{Y}$ encodes a  clique structure in the set $C$, and $X= \{x_j\}^n_{j=1}$ is the observations (radially sub-sampled frequency coefficients) in the frequency domain (\emph{k}-space). The unary potential $\psi_u$ is enforced in the \emph{k}-space while the pairwise potential $\psi_p$ is applied in the spatial domain. The unary potential enforces original observations to preserve data fidelity. Since the available observations are captured in \emph{k}-space in MRI, the model must be formulated in a way to be consistent in both \emph{k}-space and spatial domain.

The pairwise potential, on the other hand, has to be in the spatial domain to better preserve image detail since neighboring coefficients in the \emph{k}-space does not contain any meaningful spatial or data consistencies to be utilized by the pairwise potential. Therefore, the optimal way to fully utilize available data within this random field model is to formulate the unary potential in the \emph{k}-space and the pairwise potential in the spatial domain.

\begin{figure}
\begin{center}
\includegraphics[width=0.5\textwidth,keepaspectratio]{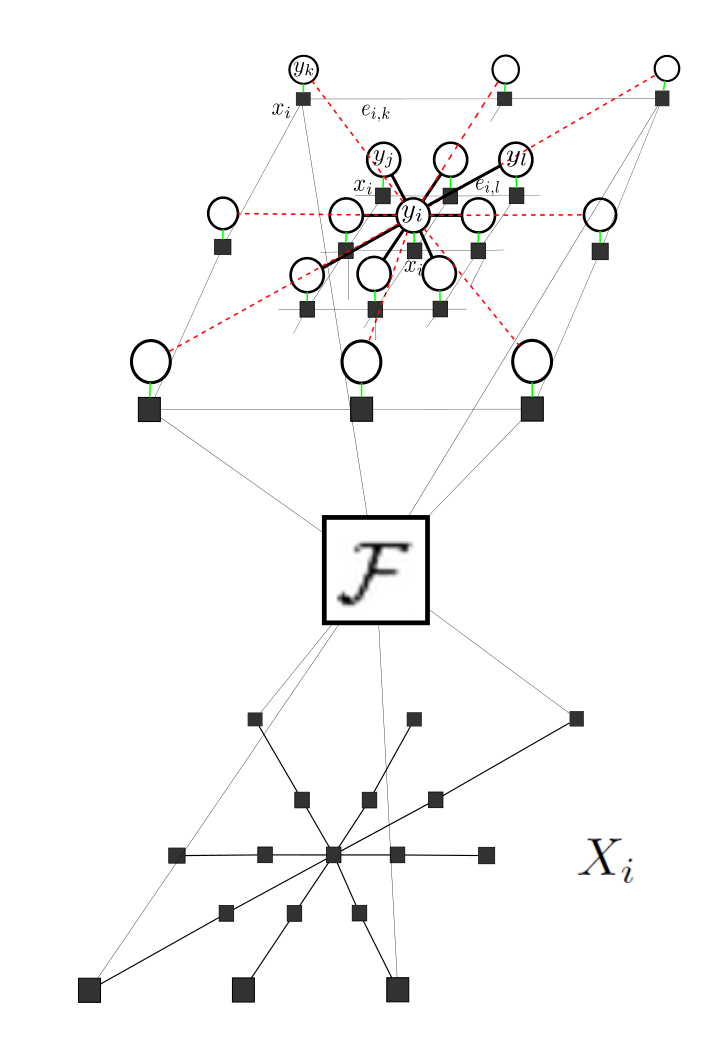}
\caption{Realization of CD-SFCRF graph.  $X_i$ represents original observations made in the \emph{k}-space, $x_i$ represents spatial domain representation of the \emph{k}-space measurements and $y_i$ represent states. \emph{F} denotes the Fourier operator used in transforming \emph{k}-space observations into the spatial domain. Connectivity is determined based on probability distributions. Nodes with higher connectivity have solid black edges while lower probable connections are represented as dashed red lines}
\label{fig:SRCRF}
\end{center}
\end{figure}

One of main differences between the proposed CD-SFCRF framework from conventional CRF models is to incorporate long-range information in the model and preserve boundaries and image structural properties more effectively which is important here due to sparse available observation.  To capture long-range information, CD-SFCRF assumes fully connected neighboring structure for the underlying graph which
each node $i$ has a set of neighbors
\begin{align}
N(i) = \biggl\{j|j=1:n,j\neq1\biggr\}
\label{NodeNeighbor}
\end{align}
\noindent where $|N(i)| = n-1$ and includes all other nodes in the graph as neighbors of node $i$. Here the pairwise clique structures are utilized such that:
\begin{align}
	C& = \biggl\{C_p(i)\biggr\}^n_{i=1}
\label{Clique}\\
C_p(i)&=\biggl\{(i,j)|j\in{N(i)},1^S_{\left\{i,j\right\}}=1\biggr\}.
\label{PWClique}
\end{align}
 The active cliques in the inference procedure are determined by  the stochastic indicator function $1^S_{\left\{i,j\right\}}=1$. The indicator function decides  whether or not  nodes can construct a clique, $C_p(i)$ for node $i$. This stochastic indicator function  combines spatial and data driven information to model the probability distribution of informative cliques which informative cliques have higher probability to participate in the inference. This combination of spatially driven and data driven probabilities can be expressed as:
\begin{align}
1^S_{\left\{i,j\right\}} = \begin{cases}
1 & P^s_{i,j}*Q^d_{i,j}\geq{\gamma}\\
0 & \text{otherwise}\\
\end{cases}
\label{neighborFunc}
\end{align}
$1^S_{\left\{i,j\right\}}=1$ brings  two properties to gather to form a clique: firstly, it incorporates the spatial information   $(P^s_{i,j})$ and secondly, it involves the data relationship between the states $(Q^d_{i,j})$ while $\gamma$ is the sparsity factor used to determine the number of active cliques in the inference. The set of active cliques are obtained to extract pairwise potentials in Eq.~\ref{PotentialEq}.
\begin{figure*}[t]
\centering
\includegraphics[width=\textwidth,keepaspectratio]{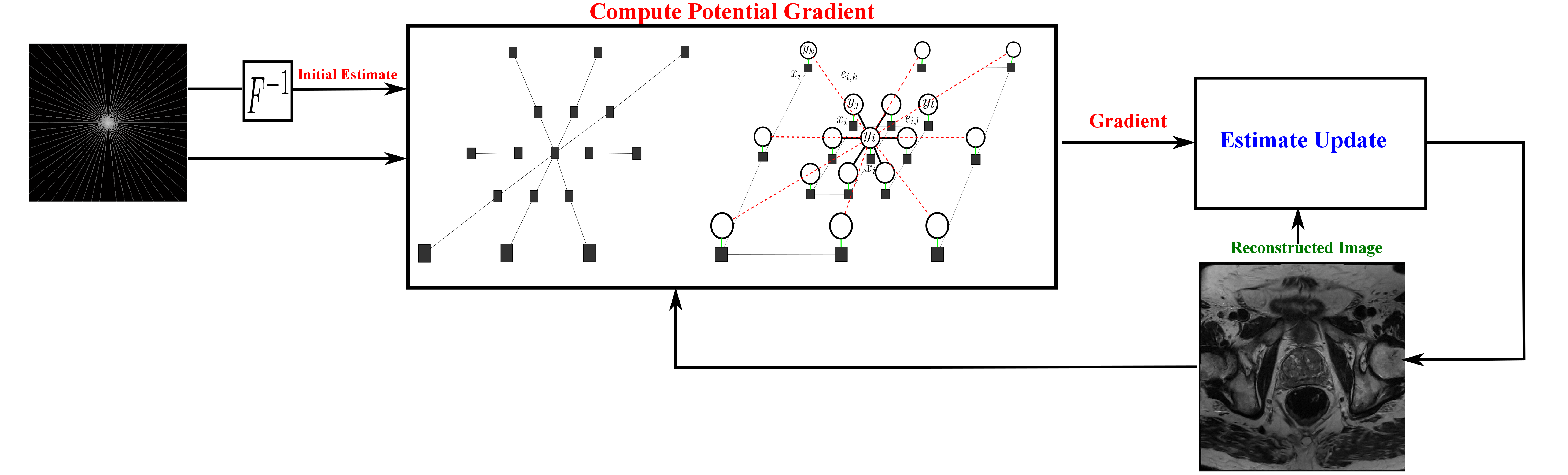}
\caption{Optimization framework of the proposed CD-SFCRF framework for sparse reconstruction from compressive sensing MRI.}
\label{fig:algorithm}
\end{figure*}
As mentioned before $\psi(\cdot)$ in Eq.~\ref{PotentialEq} is the combination of two potential functions $\psi_u(.)$, the unary potential and $\psi_p(.)$, the pairwise potential. These potential functions are formulated with their corresponding weights $\lambda$, respectively as:
\begin{align}
\psi_u(Y,X) & = \sum_{j=1}^{K}\lambda_j^uF_j(Y,X)
\label{Unary}\\
\psi_p(y_{\varphi},X) & = \sum_{\{y_i,y_j\}\in{y_{\varphi}},k=1}^{K'}\lambda_k^pf_k(y_i,y_j,X)
\label{Pairwise}
\end{align}
where  $\lambda$ controls the importance of each feature function in the energy formulation and it is calculated in the training stages. Although it is possible to provide several arbitrary feature functions to model the conditional probability $P(Y|X)$, here two feature functions are provided to formulate the image reconstruction for the purpose of sparse reconstruction from compressive sensing MRI.  The conditional distribution of $Y$ given $X$ is trained to promote/suppress different features in both the unary and pairwise potentials. Higher $\lambda_j^u$ values promotes a higher reinforcement of original observations while high $\lambda_k^p$ values promotes higher consideration of spatial and data driven neighborhood constraints. In Eq.~\ref{Unary}, $F$ refers to the frequency domain potential function. The unary potential is calculated in the \emph{k}-space while the pairwise remains in the spatial domain. This is the novelty of the CD-SFCRF and facilitates for better preservation of fine tissue details and contrast in the reconstructed image. The unary potential function $F_j(y_i,X)$ can be formulated as:
\begin{align}
F_j(Y,X) = \sum_{\omega=-\frac{\pi}{2}}^{\frac{\pi}{2}} \mathscr{F}(Y,\omega) - x_\omega
\label{CD-Unary}
\end{align}
 where $\mathscr{F}(\cdot,\cdot)$ is the Fourier operator and returns the \emph{k}-space coefficient corresponding to frequency $\omega$. Based on this formulation, the unary potential is enforced in the \mbox{\emph{k}-space} and in the inferencing step the model tries to estimate image $Y$ to be consistent to the original \emph{k}-space observation $X = \{x_\omega\}_{\omega=-\frac{\pi}{2}}^{\frac{\pi}{2}}$.

   The pairwise function $f_k(y_i,y_j,X)$ can be formulated as:
\begin{align}
f_k(y_i,y_j,X) = \exp\Big({\frac{-(y_i-y_j)^2\cdot(x_i-x_j)^2}{3\sigma^2}}\Big)
\label{CD-pairwise}
\end{align}
where $\sigma$ is a control variable for the amount of weighting node pairs in the clique $\varphi = \{i,j\}$. Contrary to the unary potential, the pairwise potential is enforced in the spatial domain.

 \subsection*{Graph Representation}

Graph $G(V,E)$ (Figure~\ref{fig:SRCRF}) is the realization of the CD-SFCRF where $V$ is the set of nodes of the graph representing states $Y=\{y_i\}^n_{i=1}$, $E$ is the set of edges in the graph. Observations $x_i\in{X}$ are made in the \emph{k}-space domain. Our final state estimations $Y$ are in the spatial domain (image). Figure~\ref{fig:SRCRF} shows the graphical representation how the spatial and \emph{k}-space domain are incorporated  to model the conditional probability $P(Y|X)$. $x_i$ comes from sparse measurements in the \emph{k}-space. In the inference procedure the \emph{k}-space observations are transformed  into the spatial domain using the Fourier transform to compute the pairwise potentials. Pairwise potentials are calculated in the spatial domain and transformed into the \emph{k}-space to combine with the unary potential and perform data fidelity. For different types of MRI data, different sparse sampling patterns can be used. Furthermore, pairwise connectivity can be trained for specific types of details and tissue structure.

The proposed CD-SFCRF framework utilizes consistencies from the spatial domain through the pairwise potential in conjunction with \emph{k}-space information through the unary potential. A combination of the two potentials is enforced simultaneously. The unary potential utilizes original observations in the \emph{k}-space, while the pairwise potential utilizes the spatial domain representation of the observation/state information and calculates pairwise potentials for nodes in the spatial domain. This allows CD-SFCRF to take advantage of the lower computation complexity introduced by the stochastically fully-connected random field model, while leveraging the original \emph{k}-space observations in improving signal fidelity.

 \subsection*{Implementation}

An implementation of the proposed CD-SFCRF framework for the purpose of sparse reconstruction from compressive sensing MRI is illustrated in Figure~\ref{fig:algorithm}.  Here, an iterative gradient descent optimization approach is employed, and can be described as follows.  First, the original compressive sensing MRI data in \emph{k}-space is transformed to the spatial domain to provide an initial estimate of the reconstructed image.  Second, the gradient of the unary and pairwise energy potentials in Eq.~\ref{Unary} and Eq.~\ref{Pairwise} is computed, where the unary data driven consistencies with respect to the original observations are enforced in the \emph{k}-space, and spatial and data driven consistencies are enforced in the spatial domain. Third, the estimate of the reconstructed image is updated based on the previous estimate and the computed gradient.  The second and third steps of this process are repeated until convergence.


\section{Experimental Setup}
To study the efficacy of the proposed CD-SFCRF method for the purpose of sparse reconstruction of compressive sensing MRI, experiments where performed involving: i) MRI data acquired of a MRI training phantom, and ii) prostate MP-MRI data of 20 patient cases.  A detailed description of the phantom data, patient data, and MRI image acquisition procedure to facilitate for the various experiments are described below.

\subsection{Phantom Data}

\begin{figure}
\centering
\includegraphics*[width = 0.4\textwidth]{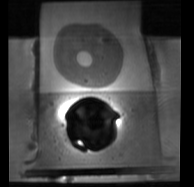}
\caption{Example slice of the prostate training phantom from Computerized Imaging Reference Systems Inc (CIRCS MODEL 053) used for evaluation purposes}
\label{fig:Phantom_Orig}
\end{figure}

The MRI training phantom used in the experiments, shown in Figure \ref{fig:Phantom_Orig}, was a multi-modality prostate training phantom from Computerized Imaging Reference Systems Inc (CIRCS MODEL 053). The phantom is composed of a clear acrylic container with dimensions $11.5 \times 7.0 \times 9.5 cm$ with a front probe opening of $3.2 cm$ diameter and a rear probe opening of $2.6 cm$ diameter. The prostate is composed of high-scattering Blue Zerdine with dimensions $5 \times 4.5 \times 4.0 cm$ and is placed in a background gel similar to water with l ittle backscatter attenuation $(\leq 0.07 dB/cm-MHz)$. Within the prostate, there are 3 randomly placed lesions of sizes between $0.5 - 1.0 cm$ placed hypoechoic to the prostate. The urethra and rectal wall are made of low scattering Zerdine with diameter of $0.7cm$ with dimensions $6 \times 11 \times 0.5 cm$ respectively. This phantom was imaged with an inflatable Medrad eCoil ERC using DWI.
The DWI MRI was acquired by a 3T GE Discovery MR750. DWI was collected at $b =  0,100,400 \text{ and } 800 s/mm^2$ at 3-NEX\textsuperscript{2} and $b=1500 s/mm^2$ 8-NEX and used to reconstruct $b=1500 s/mm^2$ collected at 16-NEX. For the DWI data, the echo time (TE) was $71.70 ms$ and repetition time (TR) was $10,000.00 ms$.

\subsection{Patient Data Experiments}

To test the efficacy of the proposed CD-SFCRF framework within a real clinical scenario, MRI data of 20 patients (17 with cancer and 3 without cancer) were acquired using a Philips Achieva 3.0T machine at Sunnybrook Health Sciences Centre, Toronto, Ontario, Canada. All data was obtained retrospectively under the local institutional research ethics board (Research Ethics Board of Sunnybrook Health Sciences Centre). For each patient, the following MP-MRI modalities were obtained (Table~\ref{t2dwi}): T2w and DWI. The patients' age ranged from 53 to 83. Table~\ref{t2dwi} summarizes the information about the 20 patients' datasets used in this study, which includes displayed field of view (DFOV), resolution, echo time (TE), and repetition time (TR).

\begin{table}[!htb]
\renewcommand{\arraystretch}{1.3}
\caption{Description of the prostate T2w and DWI images}
\label{t2dwi}
\centering
\begin{tabular}{|c|c|c|c|c|c|}
\hline
Modality & DFOV ($cm^2$) & Resolution ($mm^3$) & TE (ms) & TR (ms)\\
\hline
\hline
T2w     & $22 \times 22$  & $0.49 \times 0.49 \times 3 $& 110  & 4,687  \\
\hline
DWI     & $20\times 20$  &  $1.56 \times 1.56 \times 3$   &  61 & 6,178 \\
\hline
\end{tabular}
\end{table}
%

\begin{figure} [h!]
\centering
\includegraphics*[width = 0.4\textwidth]{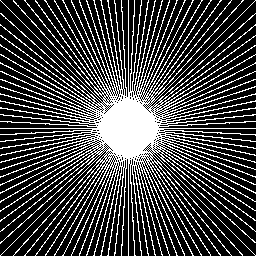}
\caption{Radial \emph{k}-space sampling pattern at $32\%$ sampling ratio.}
\label{fig:SamplingMask}
\end{figure}

\subsection{Compressed Sensing Configuration}

In order to evaluate the efficacy of the proposed CD-SFCRF framework at different sample rates, we first acquire MRI measurements at all \emph{k}-space coefficients.  Based on this fully-sampled set of \emph{k}-space measurements, sparse sampling is then conducted using radial sampling patterns with different numbers of radial sampling lines to achieve a desired sampling rate.  For example, Figure \ref{fig:SamplingMask} shows a radial sampling pattern which corresponds to a sampling rate of $32\%$ of the \emph{k}-space. Different sampling rates are tested and evaluated in this study. 

\section{Results and Discussion}
In order to evaluate the efficacy of the proposed CD-SFCRF framework for sparse reconstruction of compressive MRI sensing, a comparative evaluation analysis was performed alongside a baseline \emph{l$_{2}$} minimization (\emph{L$_{2}$}) reconstruction method, and a state-of-the-art homotopic \emph{l$_{0}$} minimization (H\emph{L$_{0}$}) \cite{trzasko2009} reconstruction method. The tested methods were compared quantitatively through peak signal-to-noise (PSNR) analysis, and qualitatively via visual assessment.  All tested methods were implemented based on the original literature, with optimal parameters used in this study. All tested methods were run until convergence. 

\begin{figure}[!ht]
\begin{center}
\includegraphics[width=0.5\textwidth]{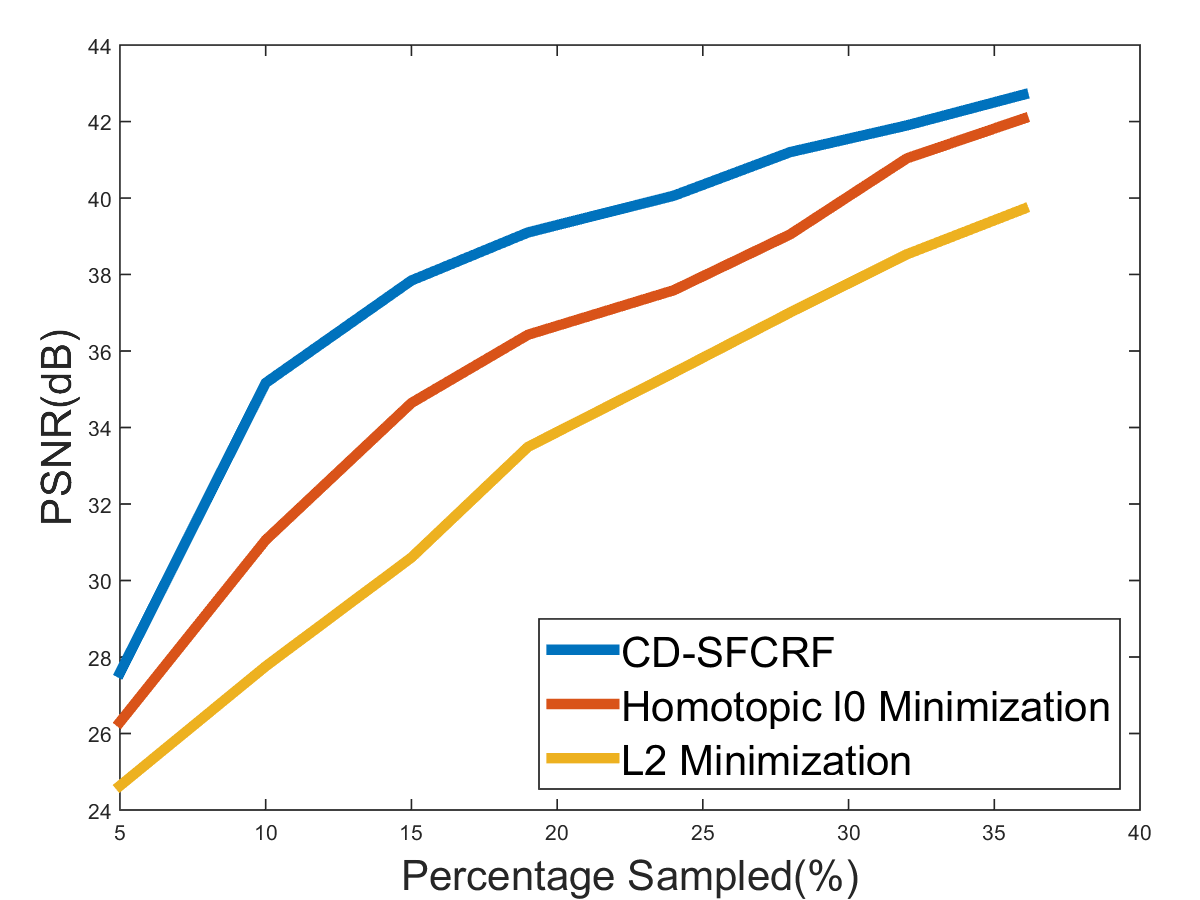}
\caption{PSNR vs. sampling percentage plots for the tested methods for the phantom MRI data at different sampling percentages.}
\label{fig:PSNR_Graph}
\end{center}
\end{figure}

Figure \ref{fig:PSNR_Graph} shows the PSNR versus sampling percentage plots for the tested methods for the phantom MRI data. It can be observed that the proposed CD-SFCRF framework achieved noticeable PSNR improvements over the other tested methods at all tested sampling percentages. The CD-SFCRF produced improvements of up to $4 dB$ over H\emph{L$_{0}$} and $7 dB$ over \emph{L$_{2}$} in low sampling conditions. It can be observed that as sampling percentage increases, the performance differences decreases. This is due to the fact that as the sampling percentage increases the amount of available measurements increases, and as such the level of reconstruction quality improvements that can be achieved will naturally decrease given the amount of available information becomes increasingly sufficient for high quality reconstruction. The ability of the CD-SFCRF framework to produce high quality reconstruction at very low sampling rates can be demonstrated visually as well.

Tables~\ref{table1},~\ref{table2}, and~\ref{table3} show the PSNR results for the three reconstructed methods for the T2w, DWI, as well as ADC map images for the patient experiments at different sampling rates. It can be observed that the proposed CD-SFCRF framework achieved the greatest PSNR improvements for the lowest sampling rate (i.e., $17\%$) where for T2w, CD-SFCRF improves PSNR by $1.78dB$ and $1.12dB$ over the \emph{L$_{2}$} and H\emph{L$_{0}$} methods, respectively. For DWI, CD-SFCRF improves PSNR by $1.85dB$ and $0.28B$ over the \emph{L$_{2}$} and H\emph{L$_{0}$} methods, respectively. Interestingly for ADC maps, the best improvements in PSNR are achieved for the highest sampling rate ($47\%$) where for CD-SFCRF improves PSNR by $4.44dB$ and $0.21B$ over the \emph{L$_{2}$} and H\emph{L$_{0}$} methods, respectively.

\begin{table}[ht!]
\begin{center}
\caption{Calculated PSNR for T2w image for the patient experiments across different methods } \label{tab:PSNR}
\label{table1}
\begin{tabular}{| c | c | c | c |}
\hline
Sampling Rate (\%) & \emph{L$_{2}$} (dB)  & H\emph{L$_{0}$} (dB) & CD-SFCRF (dB) \\
\hline \hline
  17 & 25.56 & 26.22 & \textbf{27.34}\\
  \hline
  32 & 28.39 & 28.80 & \textbf{29.72}\\
  \hline
  47& 30.42 & 30.80 & \textbf{31.23}\\
  \hline
\end{tabular}
\end{center}

\end{table}

\begin{table}[ht!]
\begin{center}
\caption{Calculated PSNR for DWI images for the patient experiments across different methods } \label{tab:PSNR}
\label{table2}
\begin{tabular}{| c | c | c | c |}
\hline
Sampling Rate (\%) & \emph{L$_{2}$} (dB)  & H\emph{L$_{0}$} (dB) & CD-SFCRF (dB) \\
\hline \hline
  17 & 26.90 & 28.46 & \textbf{28.75}\\
  \hline
  32& 31.92 & 33.39 & \textbf{33.61}\\
  \hline
  47& 36.45 & 37.85 & \textbf{37.99}\\
  \hline
\end{tabular}
\end{center}
\end{table}

\begin{table}[ht!]
\begin{center}
\caption{Calculated PSNR for ADC images for the patient experiments across different methods  } \label{tab:PSNR}
\label{table3}
\begin{tabular}{| c | c | c | c |}
\hline
Sampling Rate (\%) & \emph{L$_{2}$} (dB)  & H\emph{L$_{0}$} (dB) & CD-SFCRF (dB) \\
\hline \hline
  17 & 17.20 & 19.35 & \textbf{19.50}\\
  \hline
  32 & 18.05 & 21.66 & \textbf{21.72}\\
  \hline
  47& 18.72 & 22.94 & \textbf{23.16}\\
  \hline
\end{tabular}
\end{center}
\end{table}

Figure~\ref{fig:Results1} shows the visual comparison between between the reconstructed images produced using the proposed CD-SFCRF framework compared with that produced using the \emph{L$_{2}$} and homotopic \emph{l$_{0}$} minimization reconstruction methods for three cases for T2w images. The \emph{L$_{2}$} method resulted in blurry images as well as noticeable radial artifacts at low sampling rates. The H\emph{L$_{0}$} approach performed better than the L2 minimization and was able to noticeably reduce artifacts and provide a higher quality reconstruction.  However, in comparison, the CD-SFCRF was able to better restore details and fine tissue structure in the reconstructed image when compared to H\emph{L$_{0}$}. This is to be expected as the CD-SFCRF takes advantage of more complete data and spatial driven consistencies in a fully connected nature, thus better modeling the underlying tissue detail and structures.


\begin{figure*}[ht!]
\input{./content/bigpicture.tex}
\end{figure*}

Figures~\ref{fig:Results2} and~\ref{fig:Results3} shows the visual comparison between the reconstructed images produced using the proposed CD-SFCRF framework compared with that produced using the \emph{L$_{2}$} and H\emph{L$_{0}$} methods for three patient cases for DWI ($b=100s/mm^2$) and ADC images.  As it can be seen in both figures, the \emph{L$_{2}$} method resulted in blurry images again with noticeable radial artifacts. Although the H\emph{L$_{0}$} approach performed better than the \emph{L$_{2}$} method, it can be observed once again that the proposed CD-SFCRF approach was able to preserve more fine tissue structure and detail in the reconstructed image when compared to the H\emph{L$_{0}$} method.

\begin{figure*}[ht!]
\input{./content/dwi_results.tex}
\end{figure*}

\begin{figure*}[ht!]
\input{./content/adc_results.tex}
\end{figure*}

In Figures~\ref{fig:Results1} to~\ref{fig:Results3}, the tumourous regions marked by a radiologist and confirmed by pathology report (biopsy results) are shown by arrow and white boundary. It can be seen that the proposed CD-SFCRF method preserves the separability of the cancerous and healthy tissue in all cases, which is an important measure for usability of the proposed method in practice. As it can be seen the tumourous regions are blurred in the \emph{L$_{2}$} method, which may make it difficult to detect for radiologists.


Both quantitative and qualitative analysis demonstrate the potential of the proposed CD-SFCRF framework as a reliable reconstruction approach for compressive sensing in MRI. It demonstrates the ability to produce edge and tissue details at very low sampling rates. The CD-SFCRF framework better utilized available information to produce quality reconstruction given very limited available information. Preservation of tissue structure, detail enhancement and noise and artifact mitigation are very important for MRI as the diagnostic quality is directly related to the image quality.

Compressive sensing method used to reconstruct MR image can influence the performance of the computer-aided diagnosis (CAD) tools. For example, several radiomics-based CAD algorithms have been proposed for automatic prostate cancer detection which use T2w and DWI to extract texture and morphological features fed into a classifier~\cite{FKhalvati2015,Khalvati2014,KhalvatiMAPS,Khalvati2015_2,Khalvati2015_4,KhalvatiSuperpixel}. These algorithms heavily rely on the quality of regions of interests in similar cases in DWI and therefore, it is expected that a reconstructed MRI with better quality will improve the performance. As future work, we will investigate the effect of the proposed compressive sensing method on the detection accuracies of these radiomics-based CAD algorithms with respect to the \emph{L$_{2}$} and H\emph{L$_{0}$} methods. Moreover, recently, computational diffusion MRI (CD-MRI) has been introduced which utilizes the wealth of information in DW-MRI to computationally construct new sequences of MRI that potentially will help radiologists with more accurate and consistent diagnosis~\cite{WongCDI,Khalvati2015_1}. The proposed CD-SFCRF framework will be integrated into CD-MRI algorithms~\cite{WongCDI,Khalvati2015_1} to investigate whether CD-SFCRF improves the separability of cancerous and healthy tissues in prostate for these computationally generated MR sequences with respect to the \emph{L$_{2}$} and H\emph{L$_{0}$} methods.

\section{Conclusions}
In this study, a cross domain stochastic fully connected conditional random field (CD-SFCRF) framework for sparse reconstruction of compressive sensing MRI is presented.  The proposed CD-SFCRF framework introduces constraints in both \emph{k}-space and spatial domains within a stochastically fully connected graphical model to produce improved MRI reconstruction.  To test the efficacy of the proposed CD-SFCRF framework, quantitative experimentation using peak signal-to-noise (PSNR) analysis was performed on phantom MRI data. Quantitative and qualitative experimentation was also performed on prostate MP-MRI data of 20 patient cases at different sampling ratios. Results show an improvement over other tested sparse reconstruction approaches, especially at low sampling rates. The ability to better utilize available information given very limited information demonstrates the potential of the proposed CD-SFCRF framwork as a viable reconstruction algorithm for compressive sensing MRI.

\bibliographystyle{IEEEtran}
\bibliography{IEEEabrv,mybib2}
\end{document}